# A Sensitivity Analysis of Attention-Gated Convolutional Neural Networks for Sentence Classification

Yang Liu, Jianpeng Zhang*, Chao Gao, Jinghua Qu, Lixin Ji

National Digital Switching System Engineering and Technological R&D Center, Zhengzhou, 450002, China
Information Engineering University, Zhengzhou, 450002, China
*corresponding author's email: zjp@ndsc.com.cn

*Abstract*—In this paper, we investigate the effect of different hyperparameters as well as different combinations of hyperparameters settings on the performance of the Attention-Gated Convolutional Neural Networks (AGCNNs), e.g., the kernel window size, the number of feature maps, the keep rate of the dropout layer, and the activation function. We draw practical advice from a wide range of empirical results. Through the sensitivity analysis, we further improve the hyperparameters settings of AGCNNs. Experiments show that our proposals could achieve an average of 0.81% and 0.67% improvements on AGCNN-NLReLU-rand and AGCNN-SELU-rand, respectively; and an average of 0.47% and 0.45% improvements on AGCNN-NLReLU-static and AGCNN-SELU-static, respectively.

## I. INTRODUCTION

Natural language is inherently the unstructured data that is difficult to process and comprehension for computers. Sentence classification is one of the most essential and challenging tasks of natural language processing. Recently, Convolutional Neural Networks (CNNs) have achieved remarkable results on a number of practically important sentence classification tasks [1]-[4]. Among them, Attention-Gated Convolutional Neural Network (AGCNN) [4] improves the capability of the pooling layer in the standard CNN [1] to find the most significant features by introducing an attention gating mechanism. AGCNN is mainly constructed by two convolutional layers for different uses (see Fig. 1). AGCNN can generate the hierarchical abstract representation of the input text through the two convolutional layers. AGCNN not only allows to precisely control the length of dependencies but also enables nearby input text elements to interact at lower layers while distant text elements interact at higher layers. These characteristics make AGCNN suitable for processing the unstructured text data after the text data is mapped into word embeddings [5], [6].

The robust empirical results achieved by AGCNN demonstrate that AGCNN can be used as a substitute for traditional baseline models, e.g., statistical features-based methods [7]-[9], and standard CNNs [1]. However, AGCNN requires practitioners to set a number of hyperparameters compared to these traditional methods and the performance of AGCNN is quite sensitive to these parameters. Moreover, it is extremely costly to explore the appropriate parameter setting combination of the model in practice. Because for an AGCNN model, we need to set a large number of parameters (e.g., kernel window size of the first convolutional layer and attention-gated layer, the number of feature maps of the first convolutional layer and attention-gated layer, the dropout rate, and the activation function) and each parameter has a large value space. Also, the GPU memory usage of the model is large, and the training speed of the model is relatively slow. For example, the usage of GPU memory can reach 4377 MB, when the number of feature maps for the attention-gated layer is 10, dataset is Subj [10], and other parameters follow the original settings [4]; and running 10-fold cross-validation costs a lot of time, from about 40 minutes (CR [11] dataset) to about 2 hours (Subj dataset). These problems make the application and fine-tuning of AGCNN inconvenient.

Various emerging methods have been proposed to explore hyperparameter optimization, including random search [12], [13], Bayesian optimization [14], [15], and the combination of Bayesian optimization and Hyperband [16]. However, the vast number of possible hyperparameter configurations requires expertise to restrict the search space, and most of these sophisticated methods are very costly.

We are inspired by the previous empirical analyses of Coates et al. [17], Breuel [18], and Zhang et al. [19] on the neural networks, which explored the factors in unsupervised feature learning, the hyperparameter settings in stochastic gradient descent, and the hyperparameter settings in CNNs, respectively. Our aim in this work is to investigate and analyze the sensitivity of AGCNN to each hyperparameter setting of the architecture and to provide reasonable scope and appropriate advice for the fine-tuning of each hyper-parameter, from a large number of empirical results.

The main contributions of our work are summarized as follows:

- We investigate the sensitivity of AGCNN to each hyperparameter through a series sensitivity analysis across six different essential datasets, and we draw practical advice and reasonable scope for the tuning of each hyperparameter from a wide range of empirical results.
- We explore the effect of different combinations of hyperparameter settings on the performance of AGCNN and analyze to what extent different hyperparameter settings contribute to the performance of AGCNN.
- We improve the hyperparameter settings of AGCNN, and experiments demonstrate that our proposals achieve an average of 0.81% and 0.67%

improvements on AGCNN-NLReLU-rand and AGCNN-SELU-rand, respectively; and an average of 0.47% and 0.45% improvements on AGCNN-NLReLU-static and AGCNN-SELU-static, respectively.

The rest of the paper is organized as follows. In Section II, we review the attention mechanism and architecture of AGCNN. In Section III, we introduce the datasets and configuration of the experiment. Experimental results are summarized in Section IV. Finally, conclusions are given in Section V.

## II. RELATED WORK

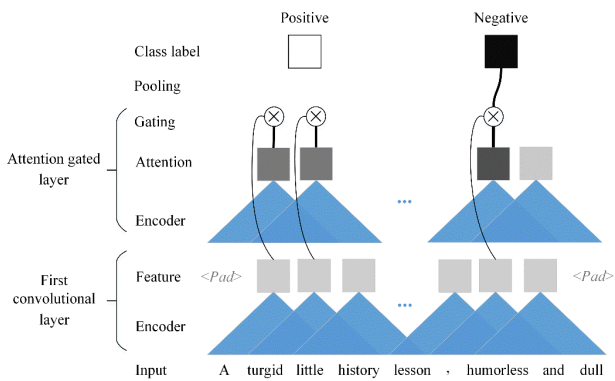

Figure 1. Illustration of the attention-gated convolutional neural network.

### A. Attention mechanism

Attention mechanism in neural networks has attracted much attention and has been applied to a variety of neural network architectures, e.g., encoder-decoder [20]. Recently, the application of the attention mechanism in CNNs has become a new research hotspot [21]. As shown in Fig. 1, the AGCNN model constructs an attention-gated layer before the pooling layer to generate attention weight from feature's context windows by using specialized convolution kernels. These specialized kernels are all one-dimensional, and their window sizes are different to obtain the different grained context attention weights of the same feature. The attention-gating mechanism on the feature maps before the pooling operation can help the pooling layer down-sample the genuinely significant abstract features.

### B. AGCNN Architecture

Fig. 1 demonstrates a simplified illustration of the AGCNNs' architecture. As depicted in Fig. 1, AGCNN consists of a convolutional layer, an attention-gated layer, a pooling layer, and a fully-connected layer with dropout operation [22] and Softmax output. The attention-gated layer contains a gating layer and a convolutional layer, where the convolutional layer (padded on the feature map when it is necessary) is used to generate attention weights.

The entire workflow of the AGCNN is as follows. First of all, each word in the input sentence is converted into a word embedding by looking up in the (pre-trained) word embedding matrix. Secondly, the abstract features are generated by the first convolutional layer and activated by activation functions. And then, through the convolution of the attention-gated convolutional layer, we can get the attention gating weights (or called the attention weights). Next, we activate these gating weights and multiply with the abstract feature maps of the first convolutional layer. Finally, we feed these attention-gated feature maps into the pooling layer and the fully connected layer to obtain the prediction output.

## III. EXPERIMENT CONFIGURATION AND DATASETS

The baseline we choose is the AGCNN-static model (with a single channel and 'static' word embeddings) [4]. For the hyperparameter settings, we set all the kernel window size to 3, and the number of feature maps is 100 and 1 for each convolutional layer, respectively. The keep rate is 0.5. Other experiment settings follow the original settings [4]. For consistency, we use the same data preprocessing steps for all the datasets as described in previous work [1], [4]. All the reported results in this paper are from 10-fold cross validation over all datasets. All experiments run with TensorFlow [23] on two NVIDIA Tesla M40 GPUs.

For investigating each parameter's effect, we hold all other settings fixed as baseline model settings and vary only the component of interest. For each configuration, we replicate the experiment 10 times and take the average as the final result. For the experiment results which are plotted as line plots, we only show the percentage change in accuracy from an arbitrary baseline model point. Since the original model has different versions based on two activation functions, i.e., Scaled Exponential Linear Unit (SELU) [24], and Natural Logarithm rescaled Rectified Linear Unit (NLReLU) [4], the model is analyzed separately based on the two activation functions. We emphasize here that in this paper, our aim is not to improve the state-of-the-art results, although the experiments show that our proposals can improve the performance of AGCNN. We aim at analyzing the sensitivity of the model on the hyperparameters and analyzing how much each setting contributes to the performance of the model.

We use six essential datasets, including one topic classification dataset (TREC [25]), one subjective/objective classification (Subj [10]), and four positive/negative classification datasets (CR [11], MR [26], SST-1 and SST-2 [27]). The data preprocessing steps we use are consistent with previous work [1], [4]. These datasets are briefly summarized as follows: (1) CR: Customer reviews of various products. (2) MR: Movie reviews dataset. (3) Subj: The snippets of movie reviews and plot summaries for movies from the internet. (4) SST-1: Extension of MR but with train/dev/test splits provided and fine-grained labels. (5) SST-2: This dataset is derived from SST-1 but removes neutral reviews and converts to two labels. (6) TREC: This dataset requires classifying questions into six question types (whether the question is about the person, location, numeric information).

Table 1 summarizes the classification accuracy of AGCNN-static model (with the baseline setting in this paper) on the six essential datasets.

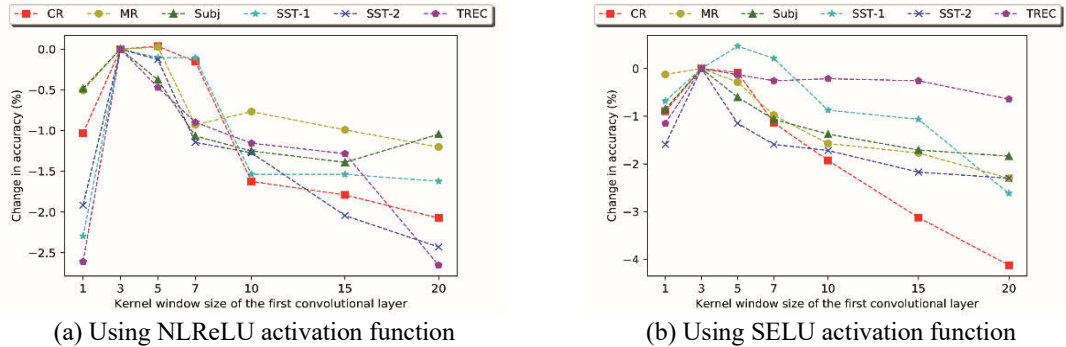

(a) Using NLReLU activation function  (b) Using SELU activation function
Figure 2. Effect of kernel window size of the first convolutional layer.

Table 1 Classification accuracy (%) of baseline on different datasets.

| Model | CR | MR | Subj | SST-1 | SST-2 | TREC |
|---|---|---|---|---|---|---|
| AGCNN-NLReLU | 84.34±0.89 | 80.72±0.85 | 92.78±0.37 | 47.47±0.97 | 85.60±0.43 | 93.48±1.21 |
| AGCNN-SELU | 85.18±0.73 | 80.83±0.69 | 93.15±0.43 | 47.02±0.72 | 86.15±0.44 | 93.68±1.02 |

## IV. SENSITIVITY ANALYSIS

### A. Effect of kernel window size of the first convolutional layer

Fig. 2 (a) and 2 (b) show the influences (on the change in the percentage of accuracy) of the kernel window size for the first convolutional layer (the kernel window size of the baseline point is 3). As shown in Fig. 2 (a) and 2 (b), as the size of the kernel window increases, the accuracy of the model will decrease significantly, which is especially apparent when the model's activation function is SELU. For the reasons that the model performance is degraded when the kernel window size is too large or too small, we believe that too large a context window cannot extract important fine-grained semantic information; similarly, too small a context window cannot extract coarse-grained semantic information. Therefore, it is necessary to combine convolution kernels of multiple different context window sizes to extract the semantic information of different granularities from the input sentences at the same time.

We then explore the case where the first convolutional layer has multiple kernel window sizes. Based on previous discussions, we permutate and combine several different window sizes from small to large. The results are reported in Table 2 and 3. It can be seen that as the combined window sizes become larger, the performance of the model decreases; and when the combination adds small window sizes, the performance rises again, and the model performs best at (1,2,3,4,5). Besides, it can be seen from the error bars of the classification accuracy in Table 2 and 3 that as the granularity of the window size combination becomes more abundant, the fluctuation of the classification accuracy of the model becomes smaller.

Therefore, the richer the granularity of the window size combination helps to improve the performance of the model.

The contextual window size of the first convolutional layer's kernels directly decides the n-gram word embedding information that AGCNNs is capable of capturing from the sentence by the convolution kernels. The combination of multiple consecutive window sizes can make the model capture different granularities information from sentences and improve the performance of the model.

### B. Effect of kernel window size of the attention-gated layer

As shown in Fig. 3 (a) and 3 (b), the effect of the kernel window size of the attention-gated layer varies between different datasets. It is noted that the performance change of the model on most datasets is more significantly as the kernel window size increases when the activation function is SELU. Comparing Fig. 2 and Fig. 3, the fluctuation of the performance caused by the change of window size in the attention-gated layer is more obvious. Therefore, AGCNN is more sensitive to window size changes in the first convolutional layer.

The performance of the model can also be improved by optimizing the window size combination of the attention-gated layer. We then explore the case where the attention-gated layer has multiple kernel window sizes (set the combination of the first convolutional layer's kernel window size to (1,2,3,4,5)). As reported in Table 4, through the grid-search we find the combination (1,3,5,7) performs the best. When the kernel window size is odd, the attention weight for each target feature is obtained from its symmetric context window; while for the even window sizes, the attention weight is obtained from the target feature's asymmetric context window [4]. Experiments show that odd window sizes can make the model perform better, since the odd-numbered attention-extracted windows are symmetric, which is beneficial to the generation of attention weights.

Table 2 Effect of multiple kernel window sizes of the first convolutional layer (using NLReLU, on the CR dataset).

| Multiple Kernel Window Sizes | Accuracy (%) |
| --- | --- |
| (1,2,3) | 85.70±0.54 |
| (2,3,4) | 85.83±0.28 |
| (3,4,5) | 85.37±0.15 |
| (4,5,6) | 85.21±0.14 |
| (1,2,3,4) | 85.63±0.27 |
| (2,3,4,5) | 85.53±0.39 |
| (3,4,5,6) | 85.29±0.27 |
| (1,2,3,4,5) | **85.92±0.38** |
| (2,3,4,5,6) | 85.37±0.36 |
| (1,2,3,4,5,6) | 85.53±0.44 |

Table 3 Effect of multiple kernel window sizes of the first convolutional layer (using SELU, on the SST-1 dataset).

| Multiple Kernel Window Sizes | Accuracy (%) |
| --- | --- |
| (1,2,3) | 47.50±0.48 |
| (2,3,4) | 47.66±0.34 |
| (3,4,5) | 47.48±0.30 |
| (4,5,6) | 47.23±0.40 |
| (1,2,3,4) | 47.60±0.38 |
| (2,3,4,5) | 47.81±0.11 |
| (3,4,5,6) | 47.34±0.21 |
| (1,2,3,4,5) | **47.90±0.29** |
| (2,3,4,5,6) | 47.45±0.27 |
| (1,2,3,4,5,6) | 47.75±0.34 |

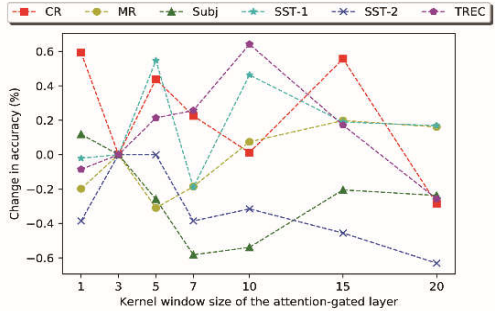

(a) Using NLReLU activation function

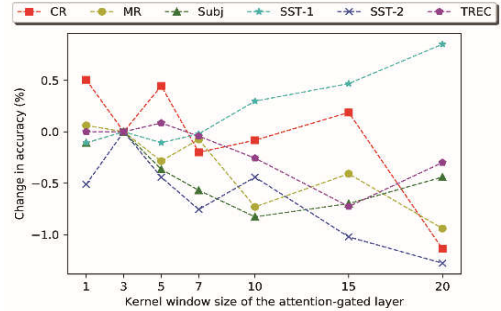

(b) Using SELU activation function

Figure 3. Effect of kernel window size of the attention-gated layer.

C. Effect of number of feature maps of the first convolutional layer

As shown in Fig. 4 (a) and 4 (b), the accuracy of the models rises rapidly as the number of feature maps in the range of 10-100, and rises slowly after 100, and tends to be stable in the range of 200-600. The increase in the number of feature maps increases the number of corresponding convolutional kernels with the same window size. Meanwhile, the increase in the number of feature maps also leads to a significant increase in the amount of model parameters and reducing the efficiency of model operation.

The choice of the number of feature maps should consider the memory allocation and model's performance comprehensively. Experimental results show that increasing the number of convolution kernels within a certain range can help the convolutional layer abstract more efficient and rich features, thus improving the performance of the model. Beyond this range, the increase of accuracy brought about by the addition of convolution kernels is limited. As illustrated in Fig. 4, 200 or more can be a good choice for the number of feature maps of the first convolutional layer. For the tuning of number of feature maps, 100 to 400 is an appropriate scope for the first convolutional layer.

D. Effect of number of feature maps of the attention-gated layer

As shown in Fig. 5 (a) and 5 (b), as the number of feature maps of the attention-gated layer becomes larger, the performance of the model on most datasets increases first and then decreases. The performance of the model on most datasets do not fall below the baseline. The percentage change in performance of the model when using SELU is more notable than that of the model when using NLReLU. Comparing Fig. 4 and Fig. 5, AGCNN is more sensitive to the number of feature maps of the first convolutional layer. Fine-tuning the number of feature maps of the attention-gated layer within a certain range can improve the performance of the model.

The convolution kernels of the attention-gated layer operate directly on the abstract features which are the n-gram word embedding-based abstract features extracted from the

Table 4 Effect of multiple kernel window sizes of the attention-gated layer (using NLReLU, on the CR dataset).

| Multiple Kernel Window Sizes | Accuracy (%) |
| --- | --- |
| (1,2,3) | 85.60±0.14 |
| (1,2,3,4) | 85.67±0.28 |
| (1,2,3,4,5) | 85.87±0.11 |
| (2,3,4) | 85.61±0.35 |
| (3,4,5) | 85.84±0.28 |
| (2,3,4,5) | 85.51±0.37 |
| (1,3,5) | 85.89±0.20 |
| (1,3,5,7) | **86.15±0.36** |
| (2,4,6) | 85.76±0.32 |
| (2,4,6,8) | 85.54±0.33 |

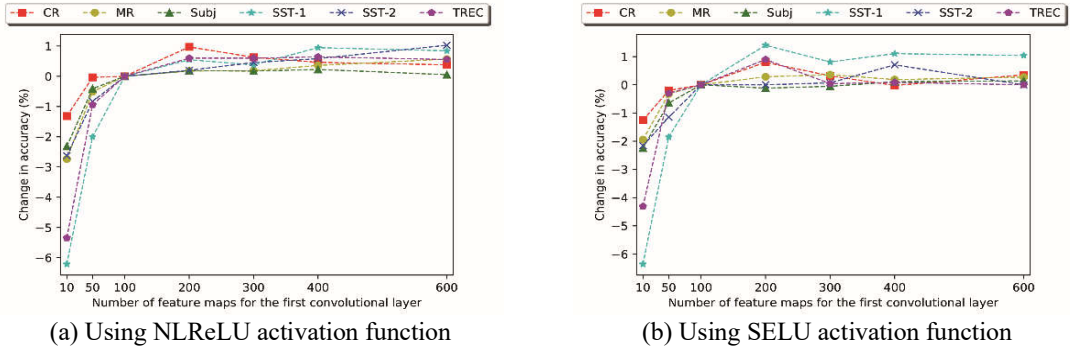

(a) Using NLReLU activation function  (b) Using SELU activation function
Figure 4. Effect of number of feature maps of the first convolutional layer.

input text. Increasing the number of feature maps within a certain range, which increases the number of kernels corresponding to the same context window size, can help the model extract more efficient and varied attention weights, thereby helping to improve the performance of the model. However, an increase of the number of feature maps of the attention-gated layer also results in a significant increase in the amount of the model's parameters and the memory usage of GPU. Therefore, a comprehensive consideration should be given to setting the appropriate number of feature maps. The number of feature maps of the attention-gated layer should be fine-tuned in the range of 10 to 50 when the GPU memory allocation and computing power are sufficient.

E. Effect of keep rate of the dropout layer

Dropout [22] is a very important regularization method for AGCNN. It is used to prevent the model from overfitting. In this section, we explore the effect of the keep rate (opposite of dropout rate) on the performance.

As shown in Fig. 6 (a) and 6 (b), a too large or too small keep rate will result in a significant drop in the model's performance. If the keep rate is too large, the model is easy to overfitting; and if the keep rate is too small, the learning of the model is insufficient, which leads to the decline in the model performance.

From Fig. 6 (a) and 6 (b), one can see that the non-zero keep rate can help with the model at some points from 0.2 to 0.8, depending on datasets. Meanwhile, it can be seen that the model is susceptible to the change of keep rate when it experiments particularly on SST-1. This indicating that the training of the model on this dataset is hard, and also the model is easy to overfit this dataset. The keep rate of the model can be fine-tuned on the particular datasets in the range of 0.2-0.8 to find the most appropriate value.

F. Effect of activation functions

Activation functions play a crucial role in achieving remarkable performance in deep neural networks. Sigmoid, ReLU [28], Softplus [28], Leaky ReLU (LReLU) [29], Parametric ReLU (PReLU) [30], Exponential Linear Unit (ELU) [31], and SELU [24] are all fairly-known and widely-used activation units.

In this section, we set ReLU as the baseline to illustrate the change percentage in the accuracy of other models using

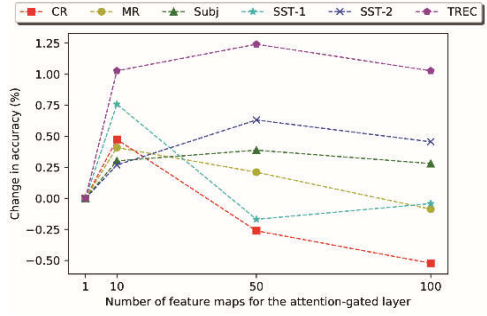 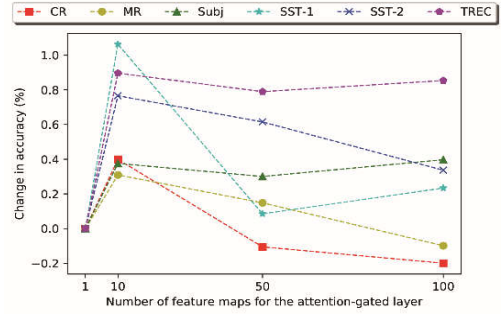

(a) Using NLReLU activation function      (b) Using SELU activation function

Figure 5. Effect of number of feature maps of the attention-gated layer.

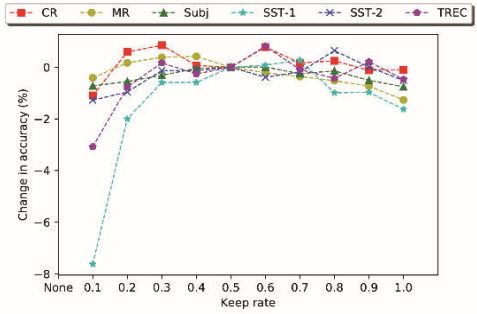 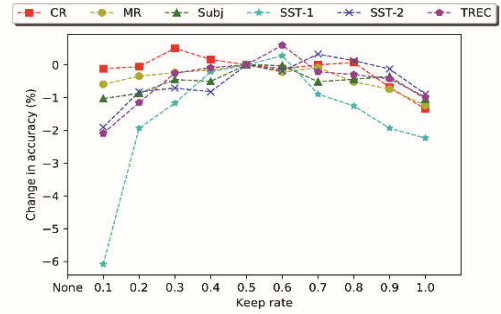

(a) Using NLReLU activation function      (b) Using SELU activation function

Figure 6. Effect of keep rate of the dropout layer.

Table 5 Effect of different hyperparameter setting combinations.

| Model | CR | MR | Subj | SST-1 | SST-2 | TREC |
|---|---|---|---|---|---|---|
| CNN-static-A | 84.74±0.89 | 81.03±0.54 | 93.04±0.34 | 45.45±0.45 | 86.65±0.21 | 92.06±0.78 |
| CNN-static-B | 85.33±0.68 | 81.14±0.96 | 93.01±0.55 | 46.60±0.30 | 86.32±0.74 | 93.21±0.59 |
| AGCNN-ReLU | 85.80±0.43 | 81.44±0.61 | 93.70±0.94 | 47.71±0.31 | 86.53±0.45 | **94.52±0.86** |
| AGCNN-NLReLU | 85.82±0.31 | 81.60±0.23 | 93.82±0.35 | 48.01±0.31 | **87.21±0.25** | 94.28±0.68 |
| AGCNN-SELU | **86.43±0.24** | **81.74±0.16** | **93.90±0.28** | **48.33±0.17** | 87.14±0.58 | 94.28±0.64 |

different activation functions compared to the baseline across different datasets.

As shown in Fig. 7, SELU is the best performing activation function, followed by NLReLU and ELU. However, ReLU performs the best on the TREC dataset. Moreover, one can see that the better performing activation functions are the ones tend to transform each layer's skewed neurons to approximately "normal" or adequately suppress the distribution of each layer's neuron activations.

The choice of activation function is essential. Although using SELU could achieve better results in most cases, it seems that AGCNNs are too sensitive to hyperparameters' changes when using SELU (compared with NLReLU). The activation function used should be determined in conjunction with the application scenario and requirements.

G. Effect of different parameter setting combinations

As shown in Table 5, in this section, we will add different parameter settings step by step on the standard CNN-static model to form different combinations of parameter settings. We aim at investigating how much these hyperparameter settings contributes to the performance of AGCNN. Models used for comparison are as follows:
- CNN-static-A: Standard CNN-static model.
- CNN-static-B: All the initializations are same as AGCNN-static but the attention-gated layer is not used, and the activation function is ReLU.
- AGCNN-ReLU(-static), AGCNN-SELU(-static) and AGCNN-NLReLU(-static): The activation functions used by each model are ReLU, SELU and NLReLU, respectively.

As reported in Table 5, from the comparison of CNN-static-A and CNN-static-B, the change of the parameter initialization method brings about an average improvement of 0.44%. The comparison between CNN-static-B and AGCNN-ReLU(-static) shows that the introducing of the attention-gated layer achieves an average performance improvement of about 0.68%. From the comparison of AGCNN-ReLU(-static), AGCNN-NLReLU(-static) and AGCNN-SELU(-static), the use of activation functions NLReLU and SELU bring about the average improvements of 0.17% and 0.35%, respectively. Therefore, the attention-gated layer contributes the most to the performance of AGCNN, followed by the initialization method and the choice of activation function.

Table 6 Classification accuracy (%) of different hyperparameter settings.

| Model | CR | MR | Subj | SST-1 | SST-2 | TREC |
|---|---|---|---|---|---|---|
| AGCNN-NLReLU-rand | 82.09±0.39 | 78.33±0.35 | 91.56±0.32 | 44.41±0.38 | 83.54±0.24 | 92.52±0.39 |
| Ours | 82.95±0.48 | 79.01±0.15 | 92.07±0.44 | 45.08±0.26 | 84.42±0.25 | 93.78±0.11 |
| AGCNN-NLReLU-static | 85.82±0.31 | 81.60±0.23 | 93.82±0.35 | 48.01±0.31 | 87.21±0.25 | 94.28±0.68 |
| Ours | 86.34±0.23 | 81.83±0.10 | 93.79±0.16 | 48.48±0.18 | 87.25±0.20 | 95.35±0.35 |
| AGCNN-SELU-rand | 82.33±0.49 | 78.29±0.31 | 91.84±0.25 | 44.72±0.18 | 83.69±0.26 | 92.93±0.41 |
| Ours | 83.02±0.43 | 78.83±0.24 | 92.27±0.15 | 45.30±0.44 | 84.35±0.12 | 94.04±0.36 |
| AGCNN-SELU-static | 86.43±0.24 | 81.74±0.16 | 93.90±0.28 | 48.33±0.17 | 87.14±0.58 | 94.28±0.64 |
| Ours | 86.67±0.19 | 81.95±0.08 | 93.72±0.12 | 48.74±0.08 | 86.90±0.33 | 95.23±0.41 |

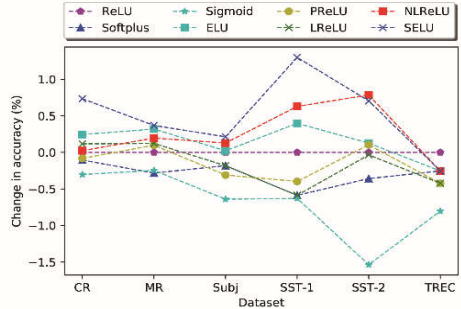

Figure 7. Effect of activation functions.

Based on the above practical results and conclusions, we improve the hyperparameter settings of AGCNN. We set the number of feature maps for the first convolutional layer and the attention-gated layer to 200 and 10, respectively. The multiple kernel window sizes for the first convolutional layer and the attention-gated layer are set to (1,2,3,4,5) and (13,5,7). Keep rate is 0.5. The results are summarized in Table 6. Compared with the baseline models, our proposals can achieve an average of 0.81% and 0.67% improvements on AGCNN-NLReLU-rand and AGCNN-SELU-rand, respectively; and an average of 0.47% and 0.45% improvements on AGCNN-NLReLU-static and AGCNN-SELU-static, respectively.

## V. CONCLUSIONS

In this paper, we investigate how sensitive the model's performance is with respect to the changes in the configurations of the parameter settings and conduct an extensive sensitivity analysis of AGCNNs for sentence classification. We then explore and analyze how much different parameter setting combinations contribute to model's performance. Meanwhile, for those interested in using AGCNNs for sentence classification in the real-world sentence classification scenarios, we draw practical advice by summarizing from these wide ranges of empirical study. Also, in this work, we improve the performance of AGCNN by improving the hyperparameter settings of AGCNN.

## ACKNOWLEDGEMENTS

We would like to thank the editors and the anonymous reviewers for their valuable comments. We acknowledge Yizhuo Yang, Tuosiyu Ming and Qiang Qu for their participations in the experiment. This work was partially supported by the Foundation for Innovative Research Groups of the National Natural Science Foundation of China (No. 61521003), and National Natural Science Foundation of China (No. 61601513, No. 61803384).